\documentclass[letterpaper]{article} 
\usepackage{aaai2026}
\usepackage{times}  
\usepackage{helvet}  
\usepackage{courier}  
\usepackage[hyphens]{url}  
\usepackage{graphicx} 
\usepackage{natbib}  
\usepackage{caption} 
\usepackage{algorithm}
\usepackage{algorithmic}
\usepackage{afterpage}
\usepackage[table]{xcolor}
\usepackage{colortbl}
\usepackage{booktabs}

\definecolor{waColor}{RGB}{200, 220, 255}  
\definecolor{wbColor}{RGB}{255, 230, 180}  

\usepackage{pifont}  

\usepackage{amsmath}    
\usepackage{amssymb}    
\usepackage{bm}         
\usepackage{makecell}
\usepackage[most]{tcolorbox}  

\definecolor{mintgreen}{RGB}{240, 255, 244}  

\usepackage{newfloat}
\usepackage{listings}
\DeclareCaptionStyle{ruled}{labelfont=normalfont,labelsep=colon,strut=off} 
\floatstyle{ruled}
\newfloat{listing}{tb}{lst}{}
\floatname{listing}{Listing}
\title{
Integrating Diverse Assignment Strategies into DETRs
}
\author{
    Yiwei Zhang\textsuperscript{\rm 1,2,5}\thanks{Work done co-mentored by Prof. Zhipeng Zhang.}, 
    Jin Gao\textsuperscript{\rm 1,2,5\text{\textdagger}}, 
    Hanshi Wang\textsuperscript{\rm 1,2,5}, 
    Fudong Ge\textsuperscript{\rm 1,2,5}, \\
    Guan Luo\textsuperscript{\rm 1,2,5},
    Weiming Hu\textsuperscript{\rm 1,2,5,6}, 
    Zhipeng Zhang\textsuperscript{\rm 3,4}\thanks{Corresponding author.}
}
\affiliations{
    \textsuperscript{\rm 1}State Key Laboratory of Multimodal Artificial Intelligence Systems (MAIS), CASIA, \\
    \textsuperscript{\rm 2}School of Artificial Intelligence, University of Chinese Academy of Sciences, \\
    \textsuperscript{\rm 3}AutoLab, School of Artificial Intelligence, Shanghai Jiao Tong University, \\
    \textsuperscript{\rm 4}Anyverse Robotics, \\
    \textsuperscript{\rm 5}Beijing Key Laboratory of Super Intelligent Security of Multi-Modal Information, \\
    \textsuperscript{\rm 6}School of Information Science and Technology, ShanghaiTech University\\

    zhipeng.zhang.cv@outlook.com, jin.gao@nlpr.ia.ac.cn, ywzhang.ai@gmail.com
}

\usepackage{bibentry}

\usepackage{multirow}
\usepackage{booktabs}

\usepackage{amsmath}
\usepackage{pifont}
\usepackage{multirow}
\usepackage{makecell}
\usepackage{enumitem}
\usepackage{graphicx}
\usepackage{subfigure}
\usepackage{subcaption}
\usepackage{float}
\usepackage{xspace}
\usepackage{colortbl}
\usepackage{diagbox}  

\def\onedot{\emph{.}}
\def\eg{\emph{e.g}\onedot}

\makeatother

\usepackage[capitalize]{cleveref}
\crefname{table}{Tab.}{Tabs.}
\Crefname{table}{Table}{Tables}
\crefname{figure}{Fig.}{Figs.}
\Crefname{figure}{Figure}{Figures}
\crefname{equation}{Eq.}{Eqs.}
\Crefname{equation}{Equation}{Equations}

\begin{document}

\maketitle
\begin{abstract}

Label assignment is a critical component in object detectors, particularly within DETR-style frameworks where the one-to-one matching strategy, despite its end-to-end elegance, suffers from slow convergence due to sparse supervision. While recent works have explored one-to-many assignments to enrich supervisory signals, they often introduce complex, architecture-specific modifications and typically focus on a single auxiliary strategy, lacking a unified and scalable design. In this paper, we first systematically investigate the effects of ``one-to-many'' supervision and reveal a surprising insight that \textit{performance gains are driven not by the sheer quantity of supervision, but by the diversity of the assignment strategies employed.} This finding suggests that a more elegant, parameter-efficient approach is attainable. Building on this insight, we propose LoRA-DETR, a flexible and lightweight framework that seamlessly integrates diverse assignment strategies into any DETR-style detector. Our method augments the primary network with multiple Low-Rank Adaptation (LoRA) branches during training, each instantiating a different one-to-many assignment rule. These branches act as auxiliary modules that inject rich, varied supervisory gradients into the main model and are discarded during inference, thus incurring no additional computational cost. This design promotes robust joint optimization while maintaining the architectural simplicity of the original detector. Extensive experiments on different baselines validate the effectiveness of our approach. 
Our work presents a new paradigm for enhancing detectors, demonstrating that diverse ``one-to-many'' supervision can be integrated to achieve state-of-the-art results without compromising model elegance. 

\end{abstract}

\begin{links}
    \link{Code}{https://github.com/Z1zyw/LoRA-DETR}
\end{links}

\section{Introduction}
\label{sec:intro}

Label assignment, which involves matching predicted objects to ground-truth annotations during training, serves as a foundational component in object detection methods. It critically determines which predictions are treated as positive or negative samples, thereby shaping the optimization process and ultimately influencing the detector's performance. Traditional detectors, such as Faster R-CNN~\cite{ren2015faster} and RetinaNet~\cite{lin2017focal}, typically employ hand-crafted rules based on anchors and IoU thresholds for this assignment, followed by Non-Maximum Suppression (NMS) to eliminate redundant predictions from overlapping anchors. In contrast, recent transformer-based frameworks like DETR~\cite{carion2020end} have introduced more principled mechanisms of one-to-one matching via the Hungarian algorithm, which assigns each object query to a unique ground-truth target. This approach effectively obviates the need for NMS during inference.

While the one-to-one assignment strategy in DETR-style detection frameworks enables a clean, end-to-end formulation, it inherently limits the number and diversity of positive samples during training, resulting in slow convergence. To address this limitation, recent studies have explored ``one-to-many'' assignment strategies to enhance supervision. However, incorporating these strategies often entails non-trivial modifications to the decoder pipeline. Such architectural modifications, as observed in certain methods~\cite{zhao2024ms_detr}, introduce unnecessary complexity that departs from the elegant simplicity of the original DETR design, making them potentially avoidable via more efficient approaches that maintain the core structure while delivering comparable or superior supervision gains. More interestingly, as we will demonstrate in the later section, the benefits of some modification appear to be less significant than previously assumed.

Moreover, in addition to this limitation, another issue is that these methods are typically tailored to a \textbf{\emph{single}} auxiliary assignment strategy, with their integration mechanisms tightly coupled to specific architectures or training procedures. This absence of a unified design hampers their scalability and flexibility, particularly when incorporating multiple assignment strategies is desired.

\begin{figure*}[t]
\centering
\subfigure[DETR]{
\begin{minipage}{0.16\linewidth}
    \centering
    \includegraphics[width=\linewidth,height=2.0in]{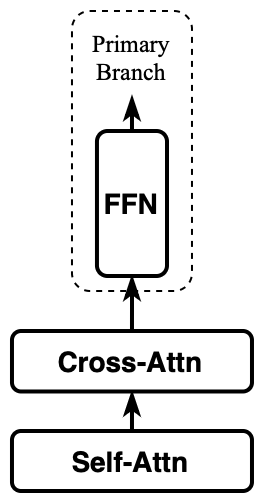}
    \label{fig:detr}
\end{minipage}}\hfill
\subfigure[MS-DETR]{
\begin{minipage}{0.21\linewidth}
    \centering
    \includegraphics[width=\linewidth,height=2.0in]{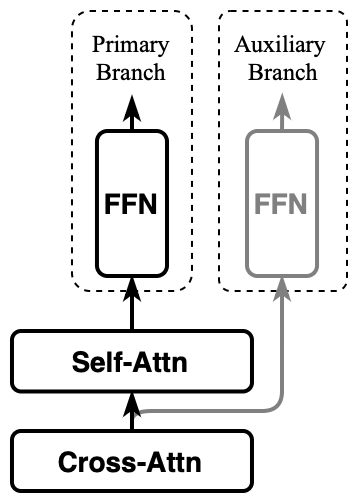}
    \label{fig:ms-detr}
\end{minipage}}\hfill
\subfigure[Multi-Branches Design]{
\begin{minipage}{0.225\linewidth}
    \centering
    \includegraphics[width=\linewidth,height=2.0in]{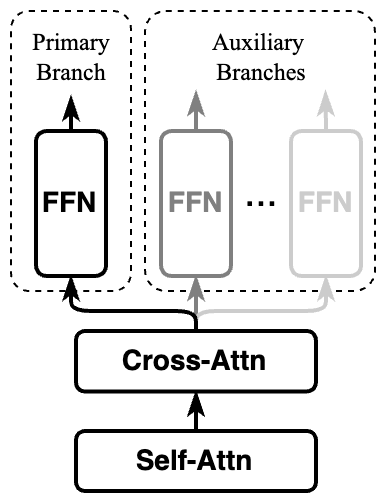}
    \label{fig:naive-multi-detr}
\end{minipage}}\hfill
\subfigure[LoRA-DETR (ours)]{
\begin{minipage}{0.20\linewidth}
    \centering
    \includegraphics[width=\linewidth,height=2.0in]{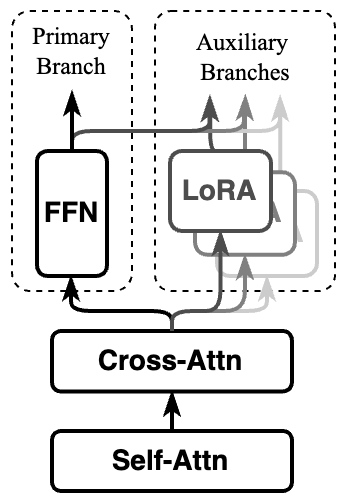}
    \label{fig:lora-detr}
\end{minipage}}
\caption{
{{Different Decoder Architectures}. (a) Standard DETR decoder layer, supervised solely by the one-to-one Hungarian matching strategy.
(b) MS-DETR design that exchanges the positions of the cross-attention and self-attention modules, enabling supervision from both one-to-one and one-to-many assignment strategies, where cross-attention is guided by the latter.
(c) A naive multi-branch extension of (a), where multiple FFNs are independently supervised by different assignment strategies.
(d) Our proposed LoRA-DETR, which integrates lightweight LoRA branches to support multiple assignment strategies with minimal structural overhead. For clarity, the figure simplifies the merge between LoRA and FFN; see~\cref{eq:lora-ffn-merge} for details. In all designs, {only the primary branch output} is fed to the next decoder layer.
}}
\label{fig:detr-arch-cmp}
\end{figure*}

Then, \textit{how can we devise a more elegant, flexible, and effective one-to-many assignment design for object detection?}

To investigate this, we begin with a straightforward design inspired by MS-DETR (\cref{fig:ms-detr}), adapting the vanilla DETR decoder layer (\cref{fig:detr}) into a multi-branch structure (\cref{fig:naive-multi-detr}). In this setup, we integrate parallel feedforward networks (FFNs) into the decoder layers and supervise each with either identical or distinct assignment strategies. The main branch and auxiliary branches respectively use one-to-one and one-to-many assignment strategies. Experimental results, detailed in \cref{tab:naive-exp}, offer valuable insights through targeted comparisons. \textbf{(1)} Simplifying MS-DETR to a DETR-like structure with an independent auxiliary FFN (\#2 \textit{vs.} \#3) results in only a negligible performance decline ($<0.1$\%), underscoring that the benefits of decoder modifications for one-to-many training have likely been exaggerated. \textbf{(2)} Building on this, experiments with additional auxiliary branches using identical one-to-many strategies (\#3, \#4, \#6) demonstrate no further gains and sometimes even degradation, revealing that sheer increases in supervision quantity can introduce counterproductive noise rather than enhancement. \textbf{(3)} In contrast, incorporating diverse assignment strategies (\#3, \#5, \#7, \#8) consistently delivers modest yet reliable improvements, suggesting a deeper mechanism at play where varied supervisory signals enrich the model's learning dynamics. Collectively, these observations emphasize supervision diversity as a pivotal driver for advancing DETR performance, often surpassing the value of mere architectural expansions in fostering robust generalization. Nevertheless, assigning dedicated FFNs to each strategy escalates parameter counts and fosters optimization redundancies, which can subtly undermine the primary branch's convergence efficiency and overall training stability.

Then, \emph{how can we introduce diverse assignment supervision in a lightweight yet more effective manner?}

\begin{table}[ht]
    \centering
    \small
    \begin{tabular}{c|ccccc}
    \hline
    \# & Model     & $N_{aux}$ & Param.     & Div.~Aux.      & mAP~$\uparrow$  \\ 
    \hline
    1   & (a)       & 0         & 47.3      &           & 43.7     \\ \specialrule{0.2pt}{0pt}{0pt}
    2   & (b)       & 1         & 53.6      &           & 47.6     \\ 
    3   & (c)       & 1         & 53.6      &           & 47.5  \\ \specialrule{0.2pt}{0pt}{0pt}
    4   & (c)       & 2         & 59.9      &       & 47.3   \\ 
    5   & (c)       & 2         & 59.9      & \checkmark          & \textbf{47.7}   \\ \specialrule{0.2pt}{0pt}{0pt}
    6   & (c)       & 3         & 66.2      &       & 47.5   \\ 
    7   & (c)       & 3         & 66.2      & \checkmark      & \textbf{48.0}   \\  \specialrule{0.2pt}{0pt}{0pt}
    8   & (c)       & 5         & 78.8      & \checkmark      & \textbf{48.2}   \\ 
    \hline
    \end{tabular}
    \caption{Results with different decoder architectures and numbers of auxiliary branches. Model correspond to \cref{fig:detr-arch-cmp}. $N_{aux}$ denotes the number of auxiliary branches. {Param.} refers to the number of parameters during training, {Div.~Aux.} indicates the use of diverse auxiliary assignment strategies, and mAP stands for mean Average Precision.}
    \label{tab:naive-exp}
    
\end{table}

To affirmatively answer the question posed by our motivation, we introduce \textbf{LoRA-DETR}, a parameter-efficient framework designed to integrate diverse assignment strategies by leveraging lightweight Low-Rank Adaptation (LoRA)~\cite{hu2022lora} branches (see~\cref{fig:lora-detr}). Specifically, we augment the primary FFN with distinct LoRA modules, each tailored to a specific assignment strategy, and jointly optimize them with the primary ``one-to-one'' branch via unified backpropagation. During the training phase, these LoRA branches serve exclusively as auxiliary paths, injecting diverse supervisory signals while maintaining maximum parameter efficiency. Crucially, they are detached during inference, ensuring that our method introduces no additional computational overhead and the inference process remains identical to the standard architecture. The LoRA mechanism facilitates the efficient modulation of the primary FFN using low-rank residual parameters, which allows the main ``one-to-one'' branch to receive gradients directly from all auxiliary tasks, thereby promoting robust parameter sharing and joint optimization. Furthermore, by adjusting the rank of each LoRA module, we can control the representational capacity of these residual parameters, thus fine-tuning the contribution of each auxiliary supervision signal to the optimization of the primary branch.

As detailed above, our paper explores the ``one-to-many'' design paradigm in object detection algorithms from a fresh perspective. Experiments reveal that simply incorporating additional branches equipped with diverse ``one-to-many'' supervision consistently enhances performance without requiring complex structural modifications. Our proposed LoRA-DETR builds on these gains and outperforms strong DETR-style baselines in various settings. For instance, with Deformable DETR~\cite{zhu2020deformable} as the baseline, our method improves its performance from 43.7 AP to 49.0 AP on COCO~\cite{lin2014coco} using three ``one-to-many'' LoRA-based auxiliary branches. Even with a strong baseline like Relation-DETR~\cite{hou2024relation}, our method boosts its performance from 51.7 AP to 52.5 AP. Within this framework, LoRA serves as an intermediary that enables more effective interactions between auxiliary tasks and the primary end-to-end detection task. Compared to simply adding auxiliary FFN networks, LoRA-DETR not only boosts parameter efficiency but also scales more effectively as the number of auxiliary branches increases.

In summary, our contributions include: \textbf{(1)} We offer a fresh perspective on the ``one-to-many'' design paradigm in object detection algorithms. Empirical results show that simply adding branches with \textbf{diverse} ``one-to-many'' supervision consistently boosts performance without complex structural changes. \textbf{(2)} We propose LoRA-DETR as an extremely simple and scalable framework that integrates diverse assignment strategies into DETR-style detectors through lightweight LoRA branches. \textbf{(3)} LoRA-DETR delivers competitive or state-of-the-art results on various baselines, highlighting the effectiveness and broad applicability of our design.

\section{Related Work}
\label{sec:related-work}

\textbf{Assignmenet Strategies in Object Detection.} 
Assignment strategies are critical in object detection. Classic two-stage detectors like Faster-RCNN~\cite{ren2015faster} use max-IoU for positive/negative assignment. Subsequent improvements, such as ATSS~\cite{zhang2020atss} introduce adaptive IoU thresholds, while one-stage detectors, such as YOLOX~\cite{ge2021yolox_simOTA} and YOLOF~\cite{chen2021yolof}, adopt SimOTA~\cite{ge2021yolox_simOTA} and top-k assignment to improve matching quality. NMS is required in these methods to resolve overlapping detection boxes.

The advent of DETR reframed detection as a set prediction task using one-to-one bipartite matching, which simplifies the pipeline but suffers from slow convergence due to sparse positive samples. To address this, recent works propose one-to-many assignment strategies to improve training efficiency. 
We categorize existing one-to-many supervision strategies into two types:
\textbf{(1)} \emph{Static Label-Query Construction}: These methods explicitly construct query-label pairs, where the supervision signal is predetermined and independent of the model’s behavior. For example, DN-DETR~\cite{li2022dn_detre} and DINO~\cite{zhang2022dino} generate positive and negative queries by perturbing ground-truth boxes, while Co-DETR~\cite{zong2023co_detr} introduces auxiliary detection heads to define additional query-label pairs.
\textbf{(2)} \emph{Dynamic Label Assignment}: These methods generate auxiliary predictions within the network and assign supervision dynamically based on matching strategies. For instance, H-DETR~\cite{jia2023h_detr} and Align-DETR~\cite{cai2023align_detr} apply one-to-many assignment strategies to intermediate decoder layers. Group-DETR~\cite{chen2023group_detr} constructs multiple groups of queries, each of which independently performs one-to-one supervision. Similarly, DAC-DETR~\cite{hu2023dac_detr}, MS-DETR~\cite{zhao2024ms_detr}, and Mr.~DETR\cite{zhang2024mr_detr} introduce parallel prediction branches and apply one-to-many supervision alongside the primary one-to-one branch.
Our method falls into the second category, but instead of relying on a \emph{single} one-to-many assignment strategy, we introduce lightweight auxiliary LoRA branches trained with diverse strategies.

\noindent \textbf{LoRA and Parameter-Efficient Adaptation.} 
Low-Rank Adaptation (LoRA)~\cite{hu2022lora} was initially introduced to efficiently fine-tune large-scale language models by injecting task-specific low-rank matrices into the weight updates. This approach enables adaptation with lightweight parameters overhead and has since been extended to various vision tasks, such as segmentation~\cite{zhang2023customized_lora_seg} and object tracking~\cite{lin2024lora_tracking} with transformers. A key advantage of LoRA is its ability to support multiple learning objectives within a shared backbone through modular and lightweight branches.
In our work, we leverage LoRA not only for parameter efficiency but also to encourage output diversity. Specifically, LoRA modules are used to decouple the learning signals from different assignment strategies, enabling the detector to produce diverse predictions.

\section{Methodology}
\label{sec:method}


\subsection{Preliminary}

\textbf{DETR Architecture.} The DETR detector comprises three main components: a backbone, a transformer encoder, and a transformer decoder. It first encodes the input image $\textbf{I}$ into a feature sequence $\textbf{X} = \{\textbf{x}_1, \textbf{x}_2, \ldots, \textbf{x}_m \}$ using the backbone and encoder. A set of object queries $\textbf{Q} = \{\textbf{q}_1, \textbf{q}_2, \ldots, \textbf{q}_n \}$, along with the image features $\textbf{X}$, are then passed to the decoder to predict classification scores $\textbf{C} = \{\textbf{c}_1, \textbf{c}_2, \ldots, \textbf{c}_n\}$ and bounding boxes $\textbf{B} = \{\textbf{b}_1, \textbf{b}_2, \ldots, \textbf{b}_n\}$. The decoder consists of multiple layers, each composed sequentially of a self-attention module, a cross-attention module, and a FFN.

\noindent\textbf{Low-Rank Adaptation.} LoRA is a lightweight fine-tuning technique that improves parameter efficiency by introducing trainable low-rank matrices into existing weight matrices of pre-trained models. Instead of updating the full weight matrix $W \in \mathbb{R}^{d \times k}$, LoRA freezes $W$ and learns a low-rank residual in the form:
\begin{align}
W'&= W + \Delta W = W + BA~,
\end{align}
where $A \in \mathbb{R}^{r \times k}$, $B \in \mathbb{R}^{d \times r}$ and $r \ll \min(d, k)$. The matrices $A$ and $B$ are the only trainable parameters, while $W$ remains frozen during training.

\noindent \textbf{Quality-Aware Loss.}
In traditional object detection, classification is treated as a binary task with hard labels ($0$ for negatives, $1$ for positives). However, this formulation does not reflect the varying quality of positive samples. Quality-aware classification loss addresses this issue by replacing hard labels with a continuous quality score~\cite{pu2023rank_detr,cai2023align_detr,zhang2021varifocalnet,huang2024deim}, allowing the model to focus more on high-quality predictions. A simple quality-aware loss, Assignment-Quality Loss, is included in experiments but is not a core part of our method.

\subsection{LoRA Branches in Decoder Layer} As showed in~\cref{fig:lora-detr}, we modified the FFN module in the decoder layers.
FFN typically consists of two linear layers $W_1$, $b_1$, $W_2$, $b_2$ with an activation function $\sigma$ (\eg ReLU or GELU) between. Formally, given an input feature $x \in \mathbb{R}^d$, the FFN computes 
\begin{align}
\text{FFN}(x) = W_2 \sigma(W_1 x + b_1) + b_2~.
\end{align}
For each auxiliary branch $i$, we insert a LoRA module into both $W_1$ and $W_2$, resulting in:
\begin{align} \label{eq:lora-ffn-merge}
\text{FFN}^{+\text{LoRA}_i}(x) & = {W'}^{i}_2 \sigma({W'}^{i}_1 x + b_1) + b_2~,
\\
{W'}^{i}_{j} & = W_{j} + B^i_j A^i_j~.
\end{align}
Unlike conventional LoRA usage where the base model is frozen to ensure parameter efficiency, we adopt LoRA as a lightweight mechanism to construct auxiliary branches. As such, both the original FFN and the LoRA modules are jointly optimized during training, allowing more effective gradient flow and interaction between the main and auxiliary objectives.
For clarity, we simplify the illustration of the merged FFN and LoRA components in \cref{fig:lora-detr}.

Given the input object queries $\textbf{Q}_j$ at the $j$-th layer, we compute the main branch output and simultaneously compute the auxiliary LoRA branch as
\begin{align}
\textbf{Q}_{j+1} &= \text{FFN} \circ \text{CA} \circ \text{SA}(\textbf{Q}_j)~, \\
\textbf{Q}_{j+1}^i &= \text{FFN}^{+\text{LoRA}_i} \circ \text{CA} \circ \text{SA}(\textbf{Q}_j)~,
\end{align}
where $\circ$ denotes the sequential composition of self-attention (SA), cross-attention (CA), and FFN.
Only the main output $\textbf{Q}_{j+1}$ is forwarded to the next layer, while the auxiliary output $\textbf{Q}_{j+1}^i$ is used solely for training. All branches within the same layer share classification head and bounding box regression head to produce detection results.

We view the LoRA module as an intermediary that absorbs the supervision signals from one-to-many assignment, while preserving the end-to-end formulation of the main one-to-one branch. This design allows us to decouple the learning objectives, enabling diverse training signals to be injected without modifying the shared query representation. 

\subsection{Assignment Strategies}

\label{subsec:assignment}

In LoRA-DETR, we apply distinct assignment strategies across decoder branches to enhance supervision diversity. The primary branch adopts the standard one-to-one matching strategy via the Hungarian algorithm, ensuring that each ground-truth box is assigned to a unique query prediction.

\begin{figure}[t]
\centering
\includegraphics[width=0.95\linewidth]{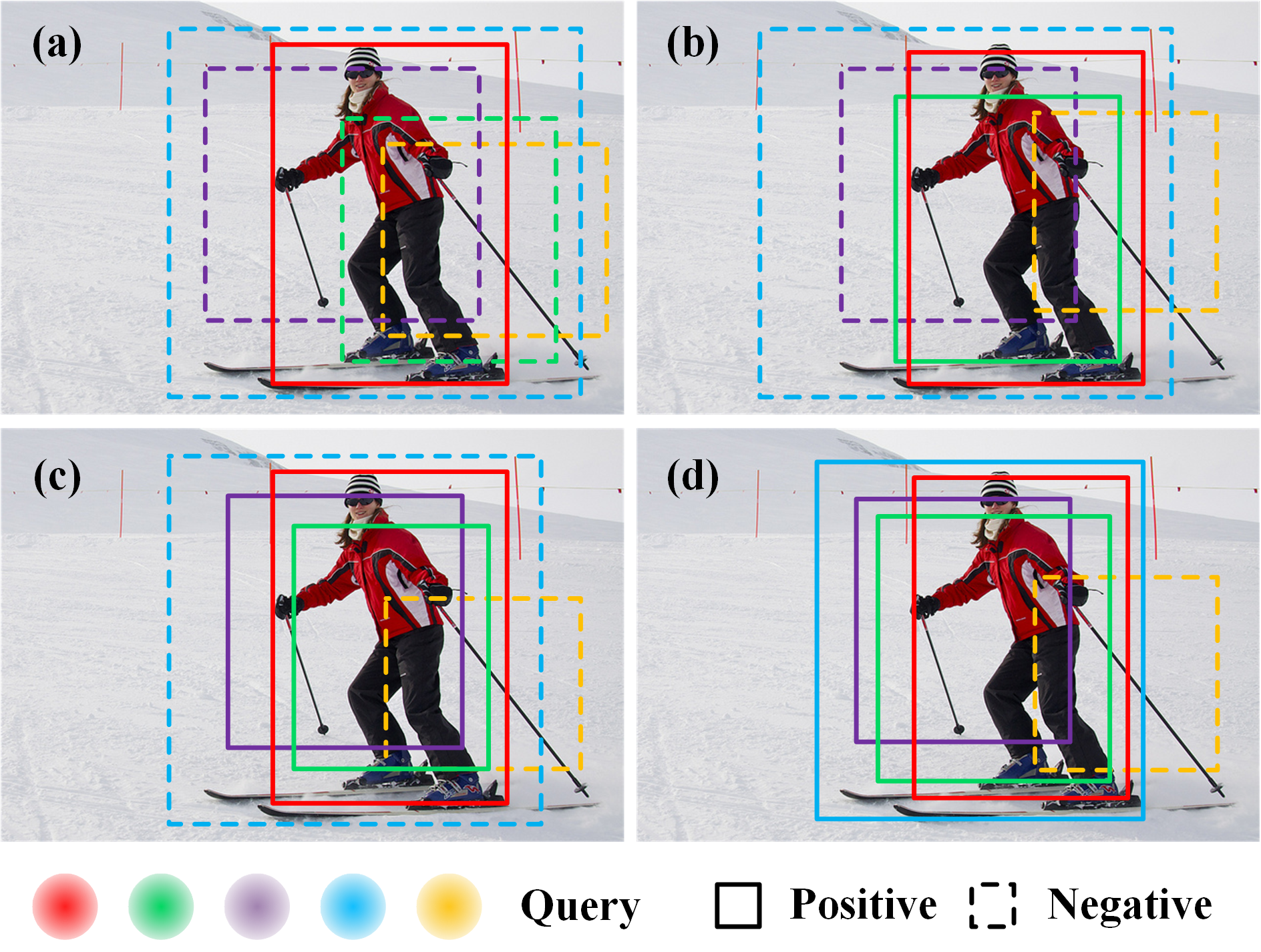}
\caption{ A set of queries simultaneously participates in diverse matching strategies. (a) illustrates the primary branch using one-to-one Hungarian matching, while (b), (c), and (d) depict auxiliary branches employing one-to-many matching with different hyperparameter of $k$.}
\label{fig:matcher-vis}
\end{figure}

For LoRA branches, we adopt one-to-many matching strategies inspired by prior works~\cite{ouyang2022nms_back, zhao2024ms_detr, hu2023dac_detr, zhang2024mr_detr}. Unlike the primary one-to-one assignment, each ground-truth box can be matched to multiple queries, providing denser supervision and accelerating convergence.
Each query prediction $\textbf{p} = <\mathbf{c}, \mathbf{b}>$ consists of a classification score $\mathbf{c}$ and a predicted bounding box $\mathbf{b}$, while a ground-truth object is denoted as $  \textbf{y} = <\hat{{c}}, \hat{\mathbf{b}}>$. A matching score between $\textbf{p}$ and $\textbf{y}$ is defined as:
\begin{align} \label{eq:quality-define}
\text{M}(\textbf{p}, \textbf{y}) = \alpha \cdot {\mathbf{c}}_{\hat{c}} + (1 - \alpha) \cdot \text{IoU}(\mathbf{b},\hat{\mathbf{b}})~.
\end{align}
Given a threshold $\tau$ and a maximum match count $k$, we select up to $k$ queries per ground-truth whose scores exceed $\tau$. These queries are treated as positives for training.
To introduce diverse supervisory signals across auxiliary branches, we vary the value of $k$ to define different one-to-many tasks. Branches with smaller $k$ focus on high-quality matches, while those with larger k allow more relaxed assignments. 

\begin{table*}[ht]
\centering
\small
\begin{tabular}{l|cc|cc|ccc}
\hline
Model             & Num. & Param. & Epoches & Queries & mAP  & AP$_{50/75}$ & AP$_{S/M/L}$ \\ \hline
Conditional-DETR~\cite{meng2021conditional_detr}  &       -             &    -                 & 50      & 300     & 40.9 & 61.8/43.3       & 20.8/44.6/60.2      \\
Anchor-DETR~\cite{wang2022anchor_detr}       &      -              &     -                & 50      & 300     & 42.1 & 63.1/44.9       & 22.3/46.2/60.0      \\
DAB-DETR~\cite{liu2022dab}          &       -             &         -            & 12      & 300     & 44.2 & 62.5/47.3       & 27.5/47.1/58.6      \\
SQR-DETR~\cite{chen2023SQR_DETR}          &      -              &       -              & 50      & 300     & 45.9 & 64.7/50.2       & 27.7/49.2/60.5      \\ \specialrule{0.2pt}{0pt}{0pt}
\#Deformable DETR   & 0                  & 47.3                & 12      & 300     & 43.7 & 62.2/46.9       & 26.4/46.4/57.9      \\
DAC-DETR           & 1                  & 47.3                & 12      & 300     & 47.1 & 64.8/51.1       & 29.2/50.6/62.4      \\
MS-DETR           & 1                  & 53.6                & 12      & 300     & 47.6 & 64.9/51.7       & 29.6/50.9/63.3      \\
\rowcolor[gray]{0.95}
LoRA-DETR         & 1                  & 48.1               & 12 & 300  & 48.8   & 65.6/52.8     & 30.4/52.6/\textbf{64.5} \\   
\rowcolor[gray]{0.95}
\rowcolor[gray]{0.95}
LoRA-DETR         & 3                  & 49.9                & 12      & 300     & \textbf{49.0} & \textbf{66.1}/\textbf{53.2}     & \textbf{31.0}/\textbf{52.8}/\textbf{64.5}    \\ \specialrule{0.2pt}{0pt}{0pt}
\#Deformable DETR++ & 0                  & 47.5                & 12      & 900     & 47.6 & 65.8/51.8       & 31.2/50.6/62.6      \\
H-DETR            & 1                  & 47.9                    & 12      & 900     & 49.0 & 66.6/53.5       & 32.0/52.3/63.0      \\
DAC-DETR          & 1                  & 47.5                    & 12      & 900     & 49.3 & 66.5/53.8       & 31.4/52.4/64.1      \\
MS-DETR           & 1                  & 53.8                & 12      & 900     & 50.0 & 67.3/54.4       & 31.6/53.2/64.0      \\
\rowcolor[gray]{0.95}
LoRA-DETR         & 1                  & 48.4                & 12      & 900     & {50.8} & {68.1}/{55.2}       & \textbf{34.3}/{54.4}/{66.0}      \\ 
\rowcolor[gray]{0.95}
LoRA-DETR         & 3                  & 51.4                & 12      & 900     &  \textbf{51.0}   & \textbf{68.5}/\textbf{55.6}       & {34.1}/\textbf{54.7}/\textbf{66.2}             \\ 
\specialrule{0.2pt}{0pt}{0pt}
\#Deformable DETR++ & 0                 & 47.5                    & 24      & 900     & 49.8 & 67.0/54.2       & 31.4/52.8/64.1      \\
MS-DETR           & 1                  & 53.8                & 24      & 900     & 50.9 & 68.4/56.1       & \textbf{34.7}/54.3/65.1      \\
\rowcolor[gray]{0.95}
LoRA-DETR         & 1                  & 48.4                & 24      & 900     &  \textbf{52.1}    & \textbf{69.4}/\textbf{56.8}           & 34.5/\textbf{55.6}/\textbf{67.0}                 \\ 
\specialrule{0.2pt}{0pt}{0pt}
\#Relation-DETR               & 1                  & 48.6                & 12      & 900     & 51.7     & 69.1/56.3  & 36.1/55.6/66.1                  \\ 
Relation-DETR               & 1                  & 48.6                & 24      & 900     & 52.1     & 69.7/56.6  & 36.1/56.0/66.5                       \\ 
MI-DETR~\cite{nan2025mi_detr}               & 1            &         $>$75.0~~~ & 12      & 900     & 52.4     & \textbf{69.8}/57.0  & 35.6/56.1/\textbf{67.2}           \\ 
\rowcolor[gray]{0.95}
LoRA-DETR          & 1+1                & 49.5                & 12      & 900     & {52.3}     & 69.4/56.9       & 36.1/56.4/66.8         \\ 
\rowcolor[gray]{0.95}
LoRA-DETR          & 1+3                & 51.3                & 12      & 900     & \textbf{52.5}     & \textbf{69.8}/\textbf{57.3}           & \textbf{37.2}/\textbf{56.5}/66.8                  \\ \hline

\end{tabular}
\caption{  {The performance on the COCO 2017 validation set.} All models use ResNet-50~\cite{he2016deep} as backbone. Num. is number of auxiliary branches and Param. is {training} parameter. We also consider the Hybrid branch proposed in H-DETR as an auxiliary branch.  \# stands for baseline. }
\label{tab:cmp-sota-coco-res50}

\end{table*}

\begin{table*}[ht]
\centering
\small
\begin{tabular}{l|cccccccc}
\hline
Model                           & Epoches & Queries & mAP  & AP$_{50}$  & AP$_{75}$ & AP$_{S}$ & AP$_{M}$ & AP$_{L}$\\ \hline
DINO                            & 12      & 900     & 56.8 & 75.4       & 62.3      & 41.1     & 60.6     & 73.5    \\
Salience-DETR~\cite{hou2024salience}    & 12      & 900     & 56.5 & 75.0   & 61.5      & 40.2     & 61.2     & 72.8    \\
Rank-DETR~\cite{pu2023rank_detr}           & 12      & 900     & 57.6 & 76.0   & 63.4      & 41.6     & 61.4     & 73.8   \\
EASE-DETR~\cite{gao2024ease}               & 12      & 900     & 57.8 & 76.7       & 63.3      & 40.7     & 61.9    & 73.7     \\
Stable-DETR                     & 12      & 900     & 57.7 & 75.7       & 63.5      & 39.8     & 62.0    & 74.7     \\
\#Relation-DETR                   & 12      & 900     & 57.8 & 76.1       & 62.9      & 41.2     & 62.1    & 74.4     \\ 
Relation-DETR                   & 24      & 900     & 58.1 & 76.3       & 63.2      & 42.0     & 62.7    & 73.7     \\ 
MI-DETR~\cite{nan2025mi_detr}                         & 12      & 900     & \textbf{58.2} & \textbf{76.5}       & \textbf{63.4}      & \textbf{42.5}     & 62.8    & \textbf{74.6}     \\
\rowcolor[gray]{0.95}
LoRA-DETR                       & 12      & 900     & \textbf{58.2} & 76.2       & \textbf{63.4}      & 41.8     & \textbf{63.0}    & 74.4               \\ 
\hline
\end{tabular}
\caption{{The performance on the COCO 2017 validation set.} All models use Swin-L~\cite{liu2021swin} as backbone. \# denotes the baseline model. We train our LoRA-DETR using \texttt{bf16} mixed-precision for improved efficiency.
}
\label{tab:cmp-sota-coco-swinL}

\end{table*}

\subsection{Assignment Quality Loss} 

For the classification loss, we adopt a simplified variant of Varifocal Loss~\cite{zhang2021varifocalnet} as our strong baseline. Following the design of DEIM~\cite{huang2024deim}, we apply the standard binary cross-entropy loss to positive samples using the quality score $s \in [0, 1]$ as the classification target, while omitting the original weighting factor introduced in Varifocal Loss. For negative samples, we retain the focal loss formulation to effectively mitigate class imbalance. The resulting loss function is defined as:
\begin{align} \label{eq:quality-aware-loss}
 & \text{VFL}^{+}(p, s, y) = \notag \\ & \begin{cases}
    -s\cdot\text{log}(p)-(1-s)\cdot\text{log}(1-p)  & \text{if}~~~y=1 \\
    - p^\gamma \cdot \text{log}(1-p)    & \text{if}~~~y=0~.
\end{cases}    
\end{align}
where $p \in [0, 1]$ denotes the predicted classification score, and $y \in \{0, 1\}$ indicates the binary label.

In our framework, the definition of the quality score $s$ varies across branches, reflecting their respective assignment strategies. Specifically, the one-to-one branch uses $s = \text{IoU}(\mathbf{b}, \hat{\mathbf{b}})$ between predicted and ground-truth boxes, while the one-to-many branches adopt the matching score $s =\text{M}(\textbf{p},\textbf{y})$ introduced in~\cref{eq:quality-define}, which combines classification confidence and localization quality. While the Quality-Aware Loss is employed in our experiments, it is not the primary focus of this paper.

\section{Experiments}
\label{sec:exp}

\subsection{Implementation Details} \label{subsec:detail}
\textbf{Dataset.}
We conduct extensive experiments on the COCO dataset~\cite{lin2014coco}, training LoRA-DETR using the \texttt{train2017} split and evaluating on \texttt{val2017}. Performance is measured using standard COCO metrics, including AP (averaged over IoU thresholds from 0.50 to 0.95 in increments of 0.05), as well as AP$_{50}$, AP$_{75}$, and scale-specific AP scores: AP$_S$, AP$_M$, and AP$_L$.
 
\noindent\textbf{Training.} Our training setup follows that of MS-DETR, including the same set of hyperparameters and encoder proposal matching strategy. For $N_{{aux}} > 1$, we average the losses across all auxiliary branches. Additionally, when denoising queries \cite{zhang2022dino} are incorporated into training, we down-weight the auxiliary losses by a factor of 0.5 to balance the overall training objectives. For the baseline trained with focal loss for classification, we adopt the assignment-quality loss proposed in~\cref{eq:quality-aware-loss} with $\gamma=1.5$. For other baselines that utilize alternative quality-aware classification losses, we follow the loss functions originally used in their respective implementations, ensuring a fair comparison. Moreover, we observed abnormally large gradient norms in the bounding box head during training and thus reduced its learning rate by a factor of 0.1.

\noindent\textbf{Assignment Strategies.} The primary branch adopts one-to-one matching, ensuring end-to-end inference. 
The LoRA branches employ the one-to-many matching strategy introduced by~\cref{eq:quality-define}. For different numbers of auxiliary branches, we use different hyperparameters $k$ in~\cref{tab:lora-num-hyper} to encourage diverse behaviors across the branches, with the corresponding results reported in~\cref{tab:naive-exp} and~\cref{tab:ablation-main}.

\begin{table}[h]
    \centering
    \small
    \begin{tabular}{l|ccccc}
    \hline
    $N_{aux}$
     & 1     & 2      & 3  & 5\\ 
    \hline
   Identical $k$  & \{6\}          & \{6,6\}          & \{6,6,6\}   
   & -   \\ 
   Diverse $k$    & \{6\}       & \{3,6\}      &  \{2,4,6\}     
   & \{2,3,4,5,6\}        \\ 
    \hline
    \end{tabular}
    \caption{
    Hyperparameter $k$ settings under different numbers of auxiliary branches ($N_{aux}$). 
    }
    \label{tab:lora-num-hyper}
    
\end{table}

\subsection{Main Results}

We integrate our approach into Deformable-DETR~\cite{zhu2020deformable} and Relation-DETR~\cite{hou2024relation}. The experimental results are presented in~\cref{tab:cmp-sota-coco-res50} and~\cref{tab:cmp-sota-coco-swinL}. Our method consistently yields stable performance improvements across different baselines. Compared to other one-to-many strategies, our approach demonstrates a clear advantage. Notably, Relation-DETR already incorporates the denoising training~(introduced in DINO) and contrastive hybrid one-to-many supervision, providing a pretty strong baseline. But surprisingly, even when directly applied to its primary branch without the contrastive setting in Relation-DETR, our method achieves a significant gain of 0.8\%. 
Moreover, with only 12 epochs of training, our approach outperforms Relation-DETR trained for 24 epochs, highlighting superior training efficiency. In contrast to MI-DETR, whose decoder introduces repeated modules and increases total inference cost (GFlops) by over 20\%~\cite{nan2025mi_detr}, our method adds no extra computation.

\subsection{Ablation Study}

\begin{figure}[ht]
\small
\centering
\subfigure[Auxiliary Number]{
\begin{minipage}{0.486\linewidth}
    \raggedright
    \includegraphics[width=\linewidth]{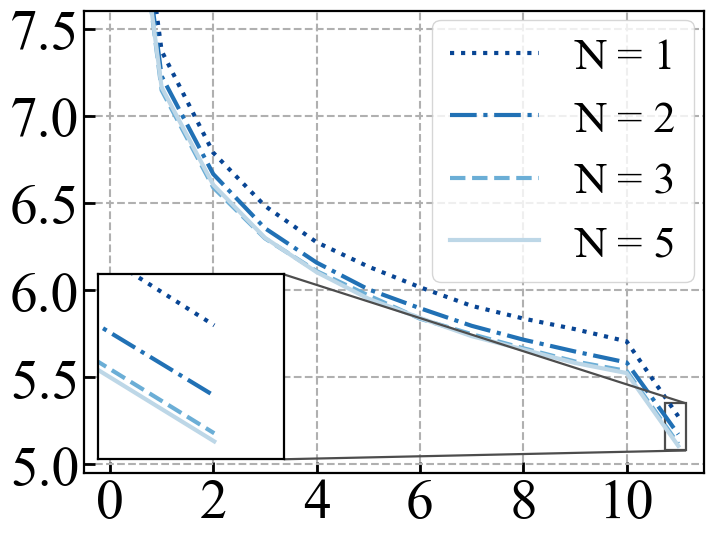}
    \label{fig:naive-num-aux-loss}
\end{minipage}}
\subfigure[Architecture]{
\begin{minipage}{0.485\linewidth}
    \centering
    \includegraphics[width=\linewidth]{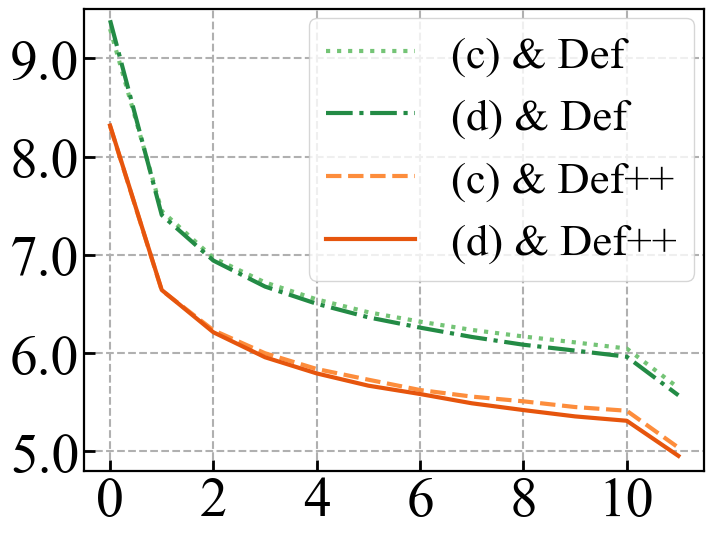}
    \label{fig:lora-vs-naive-loss}
\end{minipage}}
\caption{
{One-to-one training loss of the primary branch.} (a) Varying the number of auxiliary branches. (b) Comparing different architectural variants, as defined in ~\cref{fig:detr-arch-cmp}. The x-axis denotes epochs and the y-axis denotes the loss value. } 
\label{fig:o2o-training-loss-cmp}
\end{figure}

\noindent \textbf{Number of Auxiliary Branches.} We conduct an ablation study to investigate the impact of the number of auxiliary branches on model performance. As shown in~\cref{tab:naive-exp,tab:ablation-main}, increasing the number of auxiliary branches generally leads to consistent performance improvements, demonstrating the benefit of diverse supervision. 
As illustrated in~\cref{fig:naive-num-aux-loss}, the decrease in loss becomes relatively smaller when the number increases from 2 to 3/5. Given the trade-off between performance and computation, we choose 3 as the maximum number of auxiliary branches.

\noindent \textbf{Diverse Assignment Strategies.}
Besides varying k, diversity can also be introduced by perturbing other parameters in~\cref{eq:quality-define}. Under the 3 auxiliary branches setting, Diverse $k$ still delivers the most substantial performance gain, with 48.0 $>$ 47.7 $>$ 47.3 for Diverse $k$, Diverse $\tau$, and Diverse $\alpha$, respectively. Adjusting k thus appears to be the most effective strategy at this stage.

\begin{table}[h]
    \centering
    \small
    \begin{tabular}{c|ccccc}
    \hline
    Rank $r$    & 16      & 32   & 64  & w/o \\ 
    \hline
   Prim.        & 48.5    &  \textbf{48.8}  & 48.2 &  48.2    \\ 
   Aux. w. NMS     & 49.0          &  \textbf{49.1}     & 48.8  &   48.7  \\ 
    \hline
    \end{tabular}
    \caption{Performance~(mAP) comparison between the primary~(Prim.) and auxiliary~(Aux.) branches under different values of the hyperparameter rank $r$, where the auxiliary branch incorporates NMS. w/o means using naive FFN branch instead of LoRA.}
    \label{tab:rank-ablation}
\end{table}
\begin{table}[h]
    \centering
    \small
    \begin{tabular}{c|cc|cc}
    \hline
    Method
     & Queries    & Aux.    & Time~$\downarrow$   & Param. \\ 
    \hline
   MS-DETR      & 300          & 1$\times$ 300          & 44 &  53.8    \\ 
   DAC-DETR     & 300          & 1$\times$ 300          & 43 &  47.3    \\ 
   H-DETR       & 300          & 1$\times$300          & 46 &  47.3    \\ 
   LoRA-DETR    & 300         & 1$\times$ 300          & \textbf{40} &  48.1    \\ 
   LoRA-DETR    & 300          & 2$\times$ 300          & 46  &  49.0    \\ 
   LoRA-DETR    & 300           & 3$\times$ 300          & 53  &  49.9    \\ 
    \hline
    \end{tabular}
    \caption{{Training cost comparison with other one-to-many methods.} Aux. indicates auxiliary supervision (auxiliary branches × queries per branch). Time (minute) is GPU time per epoch on 8×A100, and Param. is training parameters.}
    \label{tab:cmp-overhead}
\end{table}
\begin{table*}[h]
    \centering
    \small
    \begin{tabular}{l|c|ccc}
    \hline
    Model                               & Param. & mAP  & AP$_{50/75}$  & AP$_{S/M/L}$      \\ \hline
    MS-DETR$*$~\cite{zhao2024ms_detr}    & 53.6   & 47.6 & 65.0/51.9     & 29.9/50.4/62.9    \\
    $+$ Architecture (b) $\to$ (c)      & 53.6   & 47.5 & 64.9/51.5     & 28.4/50.3/63.7    \\
    $+$ Box Head Learning Rate $\times$ 0.1        & 53.6   & 47.7 & 65.4/52.0     & 29.7/50.8/63.0     \\
    $+$ Assignment-Quality Loss           & 53.6   & 48.2 & 65.1/52.3     & 30.2/51.7/64.0    \\
    $+$ Architecture (c) $\to$ (d)      & 48.1   & 48.8 & 65.6/52.8     & 30.4/52.6/64.5            \\
    $+$ $N_{aux}$ $\to$  3            & 49.9   & 49.0 & 66.1/53.2     & 31.0/52.8/64.5            \\ \hline
    $+...+$ Assignment-Quality Loss $^{\dag}$           & 53.8   & 50.4 & 67.5/54.8     & 33.2/53.7/65.2    \\
    $+$ Architecture (c) $\to$ (d) $^{\dag}$     & 48.4   & 50.8 & 68.1/55.2     & 34.3/54.4/66.0            \\
    $+$ $N_{aux}$ $\to$  3 $^{\dag}$           & 51.4   & {51.0}   & {68.5}/{55.6}       & {34.1}/{54.7}/{66.2}   \\
    \hline
    \end{tabular}
    \caption{{Progressive Ablation.} 
    The table reports the model performance (AP) and corresponding parameter counts (Param.) at each stage. $*$ indicates our reproduced results. $^{\dag}$ marks results based on Deformable DETR++.}
    \label{tab:ablation-main}
\end{table*}

\noindent \textbf{LoRA Rank.} The representational capacity of the LoRA module is controlled by the hyperparameter rank $r$. A higher rank offers greater flexibility to fit the auxiliary supervision; however, if the capacity is too high, the LoRA module may overfit to the one-to-many targets or introduce conflicting gradients to the shared parameters. This can potentially slow down the convergence of the main branch or harm generalization. Conversely, too small rank may underutilize the auxiliary signals and limit the benefits of diverse supervision. 
We conduct an ablation study on the choice of LoRA rank in~\cref{tab:rank-ablation} and the best rank $r$ = 32.

\noindent \textbf{Progressive Ablation.} In~\cref{tab:ablation-main}, we present a step-by-step transition from MS-DETR to our proposed LoRA-DETR. We begin by reproducing the results of MS-DETR, obtaining a comparable mAP to the reported value, with minor variations in other metrics. We then reorder the cross-attention and self-attention modules to follow the standard sequence (Architecture (c)), which results in a slight performance drop of 0.1\%. Building on this revised architecture, we first adjust the learning rate of the bounding box head and subsequently introduce our assignment-quality loss, leading to a cumulative performance gain of 0.7\% to 48.2\%. Then, we incorporate our proposed LoRA-DETR design (Architecture (d)), which brings an additional improvement of 0.6\% on this strong baseline. Finally, as the number of auxiliary branches increases, the performance continues to improve consistently.

\noindent \textbf{Training Loss.} We present in ~\cref{fig:naive-num-aux-loss} the convergence behavior of the one-to-one training loss on the primary branch under varying numbers of auxiliary branches. As the number of auxiliary branches increases, the training loss consistently decreases across the same number of epochs, suggesting that more diverse auxiliary tasks can better facilitate the one-to-one matching process. In~\cref{fig:lora-vs-naive-loss}, comparing the convergence behavior of Arch.(c)(~\cref{fig:naive-multi-detr}) and LoRA-DETR(~\cref{fig:lora-detr}) under assignment-quality loss setting, both models perform similarly in the early training stage, but LoRA-DETR shows a consistently lower loss as training progresses. This suggests that the LoRA branch contributes more beneficial gradients to the primary branch during the later stages of training, thereby facilitating better convergence. As shown in~\cref{tab:cmp-sota-coco-res50}, under the Deformable-DETR++ setting, our method outperforms MS-DETR by 0.8\% mAP with only 12 epochs of training (50.8\% vs. 50.0\%), and achieves a larger margin of 1.2\% mAP with 24 epochs (52.1\% vs. 50.9\%), reaching performance on par with Relation-DETR~(52.1\%).

\noindent \textbf{Computation Cost Analysis.} Compared to other one-to-many approaches, our method requires only a \emph{single} forward pass through the decoder to obtain outputs from all branches, incurring only a small amount of additional computation from the LoRA modules and partially repeated FFN operations. In contrast, methods such as H-DETR and Group-DETR need to recompute the entire decoder layer for each branch. The~\cref{tab:cmp-overhead} provides a comparison of the training overhead across different methods. 

\noindent \textbf{Discussion.} MS-DETR~\cite{zhao2024ms_detr} includes an ablation study with a design similar to our naive architecture (see~\cref{fig:naive-multi-detr}), but without introducing task-specific FFNs. In this setting, the one-to-one and one-to-many branches completely share the same FFN, with separation applied only at the output heads. This shallow decoupling results in a lower mAP (47.0\%) compared to our naive design with separate FFNs (47.5\%). The performance gap highlights that, despite both branches targeting object detection, their distinct label assignment strategies lead to different optimization behaviors and should be treated as heterogeneous tasks. Completely sharing intermediate computation in such cases may cause conflicts and hinder learning. In contrast, LoRA-DETR introduces lightweight, task-specific branches using LoRA modules, enabling a controllable degree of task decoupling through rank $r$ adjustment, while maintaining minimal computational overhead.

\section{Conclusion}
\label{sec:conclusion}

This paper presents LoRA-DETR, a simple and scalable framework that introduces diverse supervision into DETR-style object detectors through lightweight LoRA-based auxiliary branches. The proposed method supports diverse assignment strategies with minimal computational and parameter overhead.
In this framework, LoRA serves as an intermediary that balances parameter sharing and task-specific specialization. 
LoRA-DETR achieves competitive or state-of-the-art results under various settings, validating the effectiveness of our design. We believe this framework provides a strong and extensible baseline for future research on efficient transformer-based detector.

\section*{Acknowledgements} This work was supported in part by the Beijing Natural Science Foundation (Grant No. JQ22014, L223003), the Natural Science Foundation of China (Grant No. 62422317, U22B2056, 62036011, 62192782, U2441241, 62503323).

\bibliography{aaai2026}

\end{document}